\documentclass[%
 aip,
 amsmath,amssymb,
 reprint,%
]{revtex4-1}

\usepackage{graphicx}
\usepackage{dcolumn}
\usepackage{bm}

\usepackage[utf8]{inputenc}
\usepackage[T1]{fontenc}
\usepackage{mathptmx}
\usepackage{etoolbox}
\usepackage{placeins}
\usepackage{siunitx}
\usepackage{rotating}
\usepackage[dvipsnames]{xcolor}

\usepackage{xcolor}

\makeatletter
\def\@email#1#2{%
 \endgroup
 \patchcmd{\titleblock@produce}
  {\frontmatter@RRAPformat}
  {\frontmatter@RRAPformat{\produce@RRAP{*#1\href{mailto:#2}{#2}}}\frontmatter@RRAPformat}
  {}{}
}%
\makeatother
\begin{document}

\preprint{AIP/123-QED}

\title[Bayesian polynomial neural ODEs]{Bayesian polynomial neural networks and polynomial neural ordinary differential equations}
\author{Colby Fronk}
\email{colbyfronk@ucsb.edu}
\affiliation{Department of Chemical Engineering, University of California, Santa Barbara, California 93106, USA}

\author{Jaewoong Yun}%
 \email{jaewoong@ucsb.edu}
\affiliation{Department of Statistics and Applied Probability, University of California, Santa Barbara, California 93106, USA}%
\affiliation{Department of Geography, University of California, Santa Barbara, California 93106, USA}

\author{Prashant Singh}%
\email{prashant.singh@scilifelab.uu.se}
\homepage{https://www.prashantsingh.se/}
\affiliation{Science for Life Laboratory, Department of Information Technology, Uppsala University, Uppsala, Sweden}

\author{Linda Petzold}
\email{petzold@ucsb.edu}
 \homepage{https://cse.cs.ucsb.edu/}
\affiliation{Department of Mechanical Engineering, University of California, Santa Barbara, California 93106, USA}
\affiliation{Department of Computer Science, University of California, Santa Barbara, California 93106, USA}

\date{\today}

\begin{abstract}
Symbolic regression with polynomial neural networks and polynomial neural ordinary differential equations (ODEs) are two recent and powerful approaches for equation recovery of many science and engineering problems.    However, these methods provide point estimates for the model parameters and are currently unable to accommodate noisy data.  We address this challenge by developing and validating the following Bayesian inference methods: the Laplace approximation, Markov Chain Monte Carlo (MCMC) sampling methods, and variational inference. We have found the Laplace approximation to be the best method for this class of problems.  Our work can be easily extended to the broader class of symbolic neural networks to which the polynomial neural network belongs.

\end{abstract}

\maketitle

\begin{quotation}
Polynomial neural ordinary differential equations (ODEs) are a recent approach for symbolic regression of dynamical systems governed by polynomials. However, they are limited in that they provide maximum likelihood point estimates of the model parameters.  The domain expert using system identification often desires a specified level of confidence or range of parameter values that best fit the data.  In this work, we use Bayesian inference to provide posterior probability distributions of the parameters in polynomial neural ODEs.  To date, there are no studies that attempt to identify the best Bayesian inference method for neural ODEs and symbolic neural ODEs. To address this need, we explore and compare three different approaches for estimating the posterior distributions of weights and biases of the polynomial neural network: the Laplace approximation, Markov Chain Monte Carlo (MCMC) sampling, and variational inference.  We have found the Laplace approximation to be the best method for this class of problems.  We have also developed lightweight JAX code to estimate posterior probability distributions using the Laplace approximation. 
\end{quotation}

\section{Introduction}

The development of a mathematical model is critical to understanding complex chemical, biological, and mechanical processes.  For example, ordinary differential equation (ODE) models are used in the field of epidemiology to describe the spread of diseases such as flu, measles, and COVID-19 and in the medical field to describe the population dynamics of CD4 T-cells in the human body during an HIV infection.  Developing a mathematical model with sufficient detail is important because it can be used to identify potential methods of intervention (such as a drug) for an undesired outcome (such as the propagation of a disease).  Scientists devote years to the model development cycle, which is the process of finding a model that describes a process, using data to fit parameters to the model, analyzing uncertainties in the fitted parameters, and performing additional experiments to refine and validate the model.  However, these mechanistic models are powerful due to their ability to directly explain the system with known first principles such as the interaction of forces, conservation of energy in the system (thermodynamics and heat transfer), and conservation of mass (transport processes).  Based on the underlying assumptions of the model, scientists know where the model can and cannot be applied to make predictions about what will happen under certain scenarios.  For these reasons, mechanistic models are preferred by scientists and engineers.  However, since these models entail a long development time, we need to develop new tools to accelerate and aid the model development cycle.  

A relatively recent development in the system identification field is the method Sparse Identification of Nonlinear Dynamics (SINDy) \cite{SINDY, BayesianSINDy, Kaheman2020SINDyPIAR}, which is linear regression of time derivatives estimated from numerical differentiation methods against a list of candidate terms which the modeler believes could be in the system to determine the terms in an ODE model.  SINDy has been shown to be very successful with recovering ODE equations from various fields including fluid dynamics \citep{PDE_SINDy}, plasma physics \citep{plasma_SINDy}, biological chemical reaction networks \citep{reaction_networks_SINDy, reactive_SINDy}, and nonlinear optical communication \citep{nonlinear_optics_SINDy}.  Like any method, SINDy is not perfect and has its flaws.  For example, it has been shown that SINDy requires its training data to be observed at very close intervals of time \cite{doi:10.1063/5.0130803}.  

The internet of things \cite{li2015internet, rose2015internet} has led to an exponential growth in the amount of data being generated and stored.  We have more data than can be effectively processed.  For example, the emergence of robots that can speedup small-scale lab experiments in chemistry and biology \citep{trends_highthroughput_screening, szymanski2011adaptation} has led to a substantially larger amount of more accurate experimental data.  In the earth sciences, the growing number of satellites and in situ earth observation equipment stationed around the world \citep{satellite} has led to a significant amount of data that must be processed and understood.  The emergence of the GPU, along with more powerful CPUs, has allowed data-driven models such as deep learning \cite{Goodfellow-et-al-2016} to emerge as a viable way to process and understand large amounts of data quickly.  

Neural ordinary differential equations \cite{chen2018neural, latent_ODEs, bayesianneuralode, stochastic_neural_ode, kidger2020neural, kidger2022neural, morrill2021neural, jia2019neural, chen2020learning, dagstuhl}  (ODEs) are a recent deep learning approach to data-driven modeling of time-series data and dynamical systems.  In Neural ODEs (NODEs), a neural network learns the right hand side of a system of ODEs.  The neural ODE is integrated forward in time from an initial condition to make a prediction. In contrast to SINDy, neural ODEs have less stringent requirements on the sampling rate, number of observed data points, and can handle irregularly spaced data points \cite{doi:10.1063/5.0130803}.  A cousin of the neural ODE is the physics-informed neural network \cite{owhadi2015bayesian, hiddenphysics, raissi2018numerical, raissi2017physics, raissi2017physics, osti_1595805, cuomo2022scientific, cai2021physics} (PINN), which attempts to accomplish the same thing but with a different approach to loss functions.

Neural differential equations and physics-informed neural networks are two powerful tools because a large majority of science and engineering models are described in terms of differential equations.  However, these tools suffer from the same major problem as the entire family of deep learning tools - they are black-box models that are not interpretable and cannot be generalized well to regimes of conditions outside of the region it was trained on.  This is an issue for scientists and engineers who need reliable models. 

In response to the need for interpretability and mechanistic models, symbolic neural networks have emerged.  There has been a recent explosion in the introduction of various symbolic neural network architectures \cite{9353253, Integration_of_Neural, Kubal_k_2023, zhang2023deep, 9659860, doi:10.1063/5.0130803, su2022kinetics, Ji_2021, 10.1063/5.0134464}, which essentially embed mathematical terms within the architecture.  Most of these architectures can be combined with neural differential equation or physics-informed neural network frameworks to recover interpretable symbolic equations \cite{doi:10.1063/5.0130803} that the scientist can immediately use.  This is referred to as symbolic regression with neural networks.

Most of these symbolic neural network approaches have been demonstrated on noiseless data only; however, real data is almost always noisy.  Additionally, the scientist using the tool often requires uncertainty estimates for the inferred model parameters; however, most of these symbolic neural network approaches recover only point estimates for the model's parameters.  Bayesian inference is one approach to handle noisy data for symbolic neural networks and symbolic neural ODEs.  There has been a substantial amount of work on Bayesian neural networks \cite{Jospin_2022} and some work on Bayesian neural ODEs \cite{bayesianneuralode, ott2023uncertainty}.  However, there is a lack of approaches attempting to find the optimal Bayesian inference method for symbolic neural networks and symbolic neural ODEs.  In our work, we explore various Bayesian inference methods and provide clarity to which Bayesian methods are best suited for this class of problems.  We evaluate the Laplace approximation, Markov Chain Monte Carlo (MCMC) sampling methods, and variational inference on our previously developed approach for symbolic regression with polynomial neural networks \cite{9353253, doi:10.1063/5.0130803} and polynomial neural ordinary differential equations \cite{doi:10.1063/5.0130803}.  Our code can easily be extended to the various other symbolic neural network architectures.

\FloatBarrier
\section{Methods}
\FloatBarrier
\subsection{Neural ODEs}
Neural Ordinary Differential Equations \cite{chen2018neural} are neural networks that learn an approximation to time-series data, $y(t)$, in the form of an ODE system.  In many fields of science, the ODE system for which we would like to learn an approximation has the form

\begin{equation}
    \frac{dy(t)}{dt} = f\left(t, y(t), \theta \right),
\end{equation}

\noindent where $t$ is time, $y(t)$ is the vector of state variables, $\theta$ is the vector of parameters, and $f$ is the ODE model.  Finding the exact system of equations for $f$ is a very difficult and time-consuming task.  With the help of the universal approximation theorem \cite{Hornik1989MultilayerFN}, a neural network ($NN$) is used to approximate the model $f$, 

\begin{equation}
    \frac{dy(t)}{dt} = f \approx NN\left(t, y(t), \theta \right).
\end{equation}

Neural ODEs can be treated like standard ODEs. Predictions for the time series data are obtained by integrating the neural ODE from an initial condition with a discretization scheme \cite{ascher1998computer, griffiths2010numerical, hairer2008solving}, in exactly the same way as it is done for a standard ODE.  

\subsection{Learning Missing Terms from an ODE Model with Neural ODEs}

When one doesn't know anything about the system's underlying equations, neural ODEs can learn the entire model:

\begin{equation}
    \frac{dy(t)}{dt} = NN\left(t, y(t), \theta \right).
\end{equation}

\noindent Often, parts of the model are known, $f_{known}$, but the modeler doesn't know all of the mechanisms and terms that describe the entire model.  In this case, we can have the neural ODE learn the missing terms:

\begin{equation}
    \label{eqn:missing_terms}
    \frac{dy(t)}{dt} = f_{known}\left(t, y(t), \theta \right) + NN\left(t, y(t), \theta \right).
\end{equation}

\noindent Learning the missing terms does not require significant special treatment, apart from including the known terms in the training process.  

\subsection{Polynomial Neural ODEs}

Systems in numerous fields are expressed as differential equations with the right-hand side functions $f$ as polynomials.  Examples include gene regulatory networks \cite{peter2020gene} and cell signaling networks \cite{gutkind2000signaling} in systems biology, chemical kinetics \cite{soustelle2013introduction}, and population models in ecology \cite{mccallum2008population} and epidemiology \cite{magal2008structured}.  Polynomial neural ODEs are useful for this class of inverse problems in which it is known a priori that the system is described by polynomials.  

Polynomial neural networks \cite{9353253, FAN2020383} are neural network architectures in which the output is a polynomial transformation of the input layer.  Polynomial neural networks belong to the larger class of symbolic neural network architectures. 

Polynomial neural Ordinary Differential Equations \cite{doi:10.1063/5.0130803} are polynomial neural networks embedded in the neural ODE framework \cite{chen2018neural}.  Since the output of a polynomial neural ODE is a direct mapping of the input in terms of tensor and Hadamard products without nonlinear activation functions, symbolic math can be used to obtain a symbolic form of the neural network.  Due to the presence of nonlinear activation functions in conventional neural networks, a symbolic equation cannot be directly obtained from conventional neural networks and conventional neural ODEs.   

\subsection{Obtaining Posterior Distributions for Weights and Biases}
We will explore and compare three different approaches for estimating the posterior distributions of weights and biases of the polynomial neural network. The approaches include the Laplace approximation, Markov Chain Monte Carlo (MCMC) sampling, and variational inference. The following text outlines each of them.

\subsubsection{Approach \#1: Laplace Approximation}

The Laplace approximation \cite{kass1991laplace} provides Gaussian approximations of the individual posteriors.  The Laplace approximation is obtained by taking the second-order Taylor expansion around the maximum a posteriori (MAP) estimate found by maximum likelihood estimation (MLE).  For the polynomial neural network, approximating the log posterior over the parameters ($\theta$), given some data ($D$) around a MAP estimate ($\theta^*$), yields a normal distribution centered around $\theta^*$ with variance equal to the inverse of the Fischer information matrix ($\mathcal{I}_\theta$):

\begin{equation}
    \theta \sim \mathcal{N}(\theta^*, \mathcal{I}_\theta^{-1}).
\end{equation}

\noindent Under certain regularity conditions, the Fisher information matrix can be calculated via either the Hessian 

\begin{equation}
\label{I_hessian_eq}
    \mathcal{I}_{\theta_{i, j}} =
  -\operatorname{E}\left[
    \frac{\partial^2}{\partial\theta_i\, \partial\theta_j} \log f(D,\theta) \right]\
\end{equation}

\noindent or the gradient 

\begin{equation}
\label{I_grad_eq}
    \mathcal{I}_{\theta_{i, j}} =
  \operatorname{E}\left[
    \left(\frac{\partial}{\partial\theta_i} \log f(D,\theta)\right)
    \left(\frac{\partial}{\partial\theta_j} \log f(D,\theta)\right)
  \right]
\end{equation}

\noindent of the log-joint density function \cite{gelman2013bayesian}.  Both the gradient and Hessian are computed with the JAX \cite{jax2018github, deepmind2020jax} automatic differentiation tool.  As expected, we were able to obtain the same results for both methods.  However, we found the calculation of the Hessian to be computationally expensive and it can only be practical for polynomial neural networks with a small number of parameters.  For this reason, we used the gradient to calculate the Fisher information.  

The log-joint density function ($\log f(D,\theta)$) is defined by the log-likelihood ($\log f(D|\theta)$) and log-prior ($\log p_r(\theta)$):

\begin{equation}
\label{eqn:logjoint}
    \log f(D,\theta) = \log f(D|\theta) + \log p_r(\theta).
\end{equation}

\noindent When the observed noise ($y_{pred} - y_{known}$) is normally distributed with variance $\beta^2$, the log-likelihood is given by: 

\begin{equation}
    \log f(D|\theta) = -\frac{1}{2 \beta^2} \sum_{i=1}^n (y_{pred} - y_{known})^2,
\end{equation}

\noindent where $y_{pred}$ is the predicted value by the polynomial neural network or polynomial neural ODE and $y_{known}$ is the observed data.  In the case of Gaussian priors on the weights and biases with covariance $\alpha^2$, the log-prior is given by:

\begin{equation}
    \log p_r(\theta) = - \frac{1}{2} \theta^T \alpha^{-2} \theta.
\end{equation}

\noindent We assume that we do not know $\beta^2$. We calculate it via the sample variance of $y_{pred} - y_{known}$ at the MAP point estimate found by MLE, with the constant term $\frac{1}{2 \beta^2}$ dropped.  Since the MLE is an unbiased estimator, $\beta^2$ can also be estimated by the mean squared error (MSE) loss \cite{mavrakakis2021probability}. 

The workflow for training Bayesian polynomial neural ODEs with the Laplace approximation is very similar to that for polynomial neural ODEs.  Prior to the training process, the architecture is defined and the parameters in the network are initialized to values that yield initial coefficient values of the simplified polynomial in the range of \num{e-5} to \num{e-10}.

The goal of the training process is to fit the neural ODE to the observed data for the state variables, $y_{known}$, as a function of time.  The neural ODE is integrated with a differentiable ODE solver to obtain predictions for $y_{known}$, which we call $y_{pred}$.  We used gradient descent \cite{lemarechal2012cauchy, hadamard1908memoire} and Adam \cite{kingma2017adam} to minimize the negative log-likelihood, with the constant term $\frac{1}{2 \beta^2}$ dropped.  

For the training process, we batch our observed data into $N_t$ batch trajectories consisting of a certain number of consecutive data points in the time series ($y_{known}$).  For each iteration (epoch) of gradient descent, we simultaneously solve $N_t$ initial value problems corresponding to each of the batch trajectories, to obtain the predictions ($y_{pred}$).  

In theory, one can use any differentiable discretization scheme to integrate the neural ODE forwards in time.  The simpler the integration scheme, the smaller the memory and compute time costs.  One can also obtain gradients for the parameters through the use of the popular continuous-time sensitivity adjoint method \cite{chen2018neural}.  Direct backpropagation through complicated integration schemes have high memory costs and numerical stability issues, therefore continuous-time sensitivity adjoint method is often used for these cases.  However, the adjoint method is very slow.  It takes a few hours to train neural ODEs with the adjoint method, whereas it only takes a few minutes to train a neural ODE with direct backpropogation through an explicit discretization scheme.  It is also important to point out that neither of these two approaches are perfect and more work needs to be done on developing differentiable ODE solvers for neural ODEs.  For example, neither direct backpropogation through an explicit scheme nor the continuous-time sensitivity adjoint method can handle obtaining gradients for stiff neural ODEs \cite{kim2021stiff}.

Since the examples we present are for non-stiff ODEs, we do not require the adjoint or any advanced integration methods, and are able to use the fourth-order explicit Runge–Kutta–Fehlberg method \cite{fehlberg1968classical} to solve the neural ODE.  The advantage of using this method is efficient direct backpropagation through the explicit ODE scheme \cite{doi:10.1063/5.0130803}, which is computationally faster than the continuous-time sensitivity adjoint method.  After the training process has converged, we have obtained the MAP estimate ($\theta^*$) via MLE.

After obtaining $\theta^*$, we can find the variance of the posterior by calculating the inverse of the Fisher Information Matrix.  For overparameterized neural network models, the Fisher Information Matrix is often singular and cannot be inverted.  In this case, an approximation to the inverse can be calculated by either the Moore–Penrose inverse \cite{ben2006generalized} or by dropping the off-diagonal entries from the matrix \cite{Cramer-Rao_Bounds}.  We have had success with both of these methods for finding an approximation for the inverse of the Fisher information.  For the case in which the matrix is invertible, the approximations have given similar results to the direct matrix inverse.  All of our results calculate the inverse using the Moore–Penrose inverse \cite{ben2006generalized}.  We have prior experience using the Laplace approximation to obtain uncertainties for the output of a neural network.  Based on our experience, the Moore–Penrose inverse can only be used on neural networks with less than 50,000 parameters.  This is because it becomes too expensive to invert the singular Fisher information matrix.  

\subsubsection{Approach \#2: Markov chain Monte Carlo}

This approach for obtaining posterior distributions for the weights and biases of the polynomial neural network draws from Markov chain Monte Carlo \cite{chen2012monte, liang2011advanced, mcelreath2018statistical} (MCMC) methods for training Bayesian neural networks \cite{9756596} (BNNs).  The two MCMC sampling methods that we explored were Hamiltonian Monte Carlo (HMC) and The No-U-Turn-Sampler (NUTS).  

Hamiltonian Monte Carlo \cite{DUANE1987216, Neal1996} (HMC) is a MCMC method that uses derivatives of the density function to generate efficient transitions.  HMC starts with an initial set of parameter values.  For a set number of iterations, a momentum vector is sampled and integrated following Hamiltonian dynamics \cite{leimkuhler2004simulating} with the leapfrog \cite{ascher1998computer} integrator with a set discretization time ($\epsilon$) and number of steps (L).  Since the leapfrog integrator incurs numerical error \cite{ascher1998computer}, it is corrected by use of the Metropolis–Hastings \cite{whitlock1986monte, tierney1994markov, 10.1093/biomet/57.1.97, 10.2307/2684568} acceptance algorithm, which helps to decide whether to accept or reject the new state predicted from Hamiltonian dynamics.  

The No-U-Turn-Sampler \cite{hoffman2014no} (NUTS) is an extension of HMC that automatically determines when the sampler should stop an iteration.  The algorithm automatically chooses the discretization time and number of steps, which avoids the need for the user to specify these additional parameters.  However, we have found this algorithm to be computationally more expensive than vanilla HMC for this class of problems.

The training process is slightly different than for the Laplace approximation.  We still batched our observed data into $N_t$ batch trajectories and simultaneously solved $N_t$ initial value problems with the same fourth-order explicit Runge–Kutta–Fehlberg method.  We used BlackJAX \cite{blackjax2020github}'s sampling algorithms to do the MCMC inference.  For both of these methods, we used the log-joint density defined in Equation \ref{eqn:logjoint}.  

\subsubsection{Approach \#3: Variational Inference}

In variational inference \cite{cinelli2021variational, nakajima2019variational, smidl2006variational}, we learn an approximation $q(\theta)$ to our posterior $p(\theta | D)$.  Our approximation is assumed to belong to a certain family of probability density functions and the parameters of that family are optimized by minimizing the Kullback–Leibler (KL) divergence:

\begin{equation}
    \text{KL}(q(\theta) \, ||\, p(\theta | D)) = \mathbb{E}_{q(\theta)} \left[\log \frac{q(\theta)}{ p(\theta | D)} \right].
\end{equation}

\noindent We don't know the analytical form of the posterior so we cannot minimize the KL divergence directly, but we can use a trick called the Evidence Lower Bound (ELBO) \cite{cinelli2021variational, nakajima2019variational, smidl2006variational}:

\begin{equation}
\label{fig:ELBO}
\text{ELBO} = \mathbb{E}_{q(\theta)}[\log p(D | \theta)] - \text{KL}(q(\theta) \, ||\, p_r(\theta)).
\end{equation}

\noindent Maximizing the ELBO is mathematically equivalent to minimizing the KL divergence.  The ELBO only contains the prior $p_r(\theta)$ and likelihood $p(D | \theta)$, which we can numerically calculate.

We wrote our own custom JAX code for variational inference.  The neural ODEs are numerically integrated exactly the same way as was done for the Laplace approximation.  We used a multivariate Gaussian distribution for the approximation $q(x)$.

\subsection{Obtaining Posterior Distributions for Polynomial Coefficients}

The polynomial neural network is a factorized form of a polynomial. To obtain a simplified form of the polynomial we must expand the equation and combine like terms.  For the case where the neural network parameters are scalar point estimates, we have already done this \cite{doi:10.1063/5.0130803} with the use of SymPy \cite{SymPy}. When our parameters are Bayesian probability distributions, we must use the rules for the product and sum of probability distributions.  These rules depend on the type of probability distributions that are algebraically combined, which makes it challenging to compute for even a small number of parameters (weights and biases).  We explored approximating the weights and biases as independent univariate Gaussian probability density functions (PDFs), for which there are known rules \cite{Bromiley2013ProductsAC} for the mean and variance of the product and sum of univariate Gaussian PDFs.  However, this approach did not work in all cases since the weights and biases are dependent on each other.  

To avoid multiplying out probability density functions of the weights and biases to obtain posterior distributions for the polynomial coefficients, we used Monte Carlo sampling.  We drew random samples from the posterior distributions $w \sim P(w|D)$ and $b \sim P(b|D)$ for the weights ($w$) and biases ($b$) given the data ($D$).  For each sample, we used the approach of expanding the polynomial neural network for scalar point estimates \cite{doi:10.1063/5.0130803}.  After doing this for enough samples, we have an estimate of the posterior distribution $c \sim P(c|D)$ of the polynomial coefficients ($c$).

\subsection{Strategies for Handling Large Amounts of Noise}

Neural ODEs require initial conditions to generate predicted trajectories ($y_{pred}$) for the training process.  When there is a large amount of observed noise in the training data, the known data points ($y_{known}$) cannot be used as initial conditions.  When this is the case, we must use a time-series filtering or smoothing algorithm to find good initial conditions to use for the neural ODE training process.  Example filtering algorithms include moving average \cite{SMITH2003277} (MA), exponential moving average \cite{6252962} (EMA), and Kalman filters \cite{kalman1961new}.  Example smoothing algorithms include smoothing splines \cite{wang2011smoothing}, local regression \cite{cleveland1996smoothing}, kernel smoother \cite{wand1994kernel}, Butterworth filter \cite{selesnick1998generalized}, and exponential smoothing \cite{vetterli2014foundations}.  We applied all of these algorithms on noisy ODE time series data and found Gaussian process regression (GPR) \cite{roberts2013gaussian} to be the most accurate approach. For brevity, we have chosen not to outline in detail the pros and cons of each of the possible algorithms.  However, it is important to note that the optimal smoothing algorithm is dependent on the data and the underlying model that describes it.  

Gaussian process regression assumes a Gaussian process prior, which is specified with mean function $m(x)$ and covariance function or kernel $k(x,x')$:

\begin{equation}
    f(x) \sim GP\left( m(x), k(x,x') \right).
\end{equation}

\noindent The rational quadratic, Matérn, Exp-Sine-Squared kernel, and radial basis function kernels \cite{genton2001classes, kocijan2016modelling, duvenaud2014automatic} were found to perform the best for our considered test problems and settings.  We used the scikit-learn \cite{scikit-learn} Python library to perform our pre-processing with GPR.  The hyperparameters of the kernels were optimized using MLE.

\section{Results}

We will start by evaluating the methodology outlined on univariate cubic regression with a polynomial neural network.  Starting with this model demonstrates that we can recover accurate Bayesian uncertainties on a standard polynomial without any ODEs.  Since this problem can be posed as a Bayesian linear regression model with a closed form solution, we can directly test the accuracy of the methods and make sure they work prior to moving on to ODEs.  

We then move on to the following ODE models: the Lotka-Volterra deterministic oscillator, the damped oscillator, and the Lorenz attractor.  These models are common toy problems for dynamical systems and neural ODEs.  The Lotka-Volterra model is a fairly easy model to identify.  The damped oscillator is more difficult.  In our previous work, we have shown that the dampening effect makes the vector field hard to learn.  Since the Lorenz attractor is chaotic and has high frequency oscillations, it is the most difficult model to learn.  Since it is common in the sciences to have a partially incomplete model, we also demonstrate learning the missing terms from a partially known ODE model.  For simplicity reasons, we have chosen to use the Lotka-Volterra model for learning the missing dynamics.   

For each of the models outlined, we recover Bayesian posterior distributions for the model parameters and compare them to the known values.  For the univariate cubic regression example, we plot the prediction along with confidence intervals and confirm that the confidence intervals capture the data well.  For the ODE examples, we integrate the Bayesian ODE models from the known initial condition and compare it to the true trajectory.  The criteria for choosing the best Bayesian inference method are: ease of use, computational cost, and accuracy.

\subsection{Univariate Cubic Regression}

Prior to studying dynamical systems with neural ODEs, we tested our Bayesian polynomial neural network inference method on basic polynomials.  For the test case, we used the following third order univariate function:

\begin{equation}
    f(x) = 1 + x + 2x^2 + 4x^3.
\end{equation}

\noindent The training data for the $x$-values consisted of 200 uniformly spaced data points in the range -1.25 to 1.25.  The values of $f(x)$ corresponding to the values of $x$ were obtained by directly substituting the $x$-values into the function.  We then added Gaussian noise with $\mu=0$ and $\sigma^2=9$ to the training data.  We chose this level of noise to demonstrate our methodology on data with a high level of noise.  For reproducibility and comparison purposes, we used a random seed of 989 for all of the results we will show.  

The architecture from Ref.~\onlinecite{doi:10.1063/5.0130803} was used for the Laplace approximation, the No-U-Turn Sampler (NUTS) method, and variational inference.  The third order polynomial neural network we used had 1x10x10x10x10x1 neurons in each layer (180 total parameters).  We experimented with changing the number of neurons in each hidden layer up to 200 and the results were similar.  The extra parameters do not affect the posteriors significantly.  For brevity, we do not show these results.  We used the Python libraries JAX \cite{jax2018github, deepmind2020jax} along with Flax \cite{flax2020github} for our neural networks.

\begin{figure}
    \centering
    \includegraphics[width=1.0\textwidth]{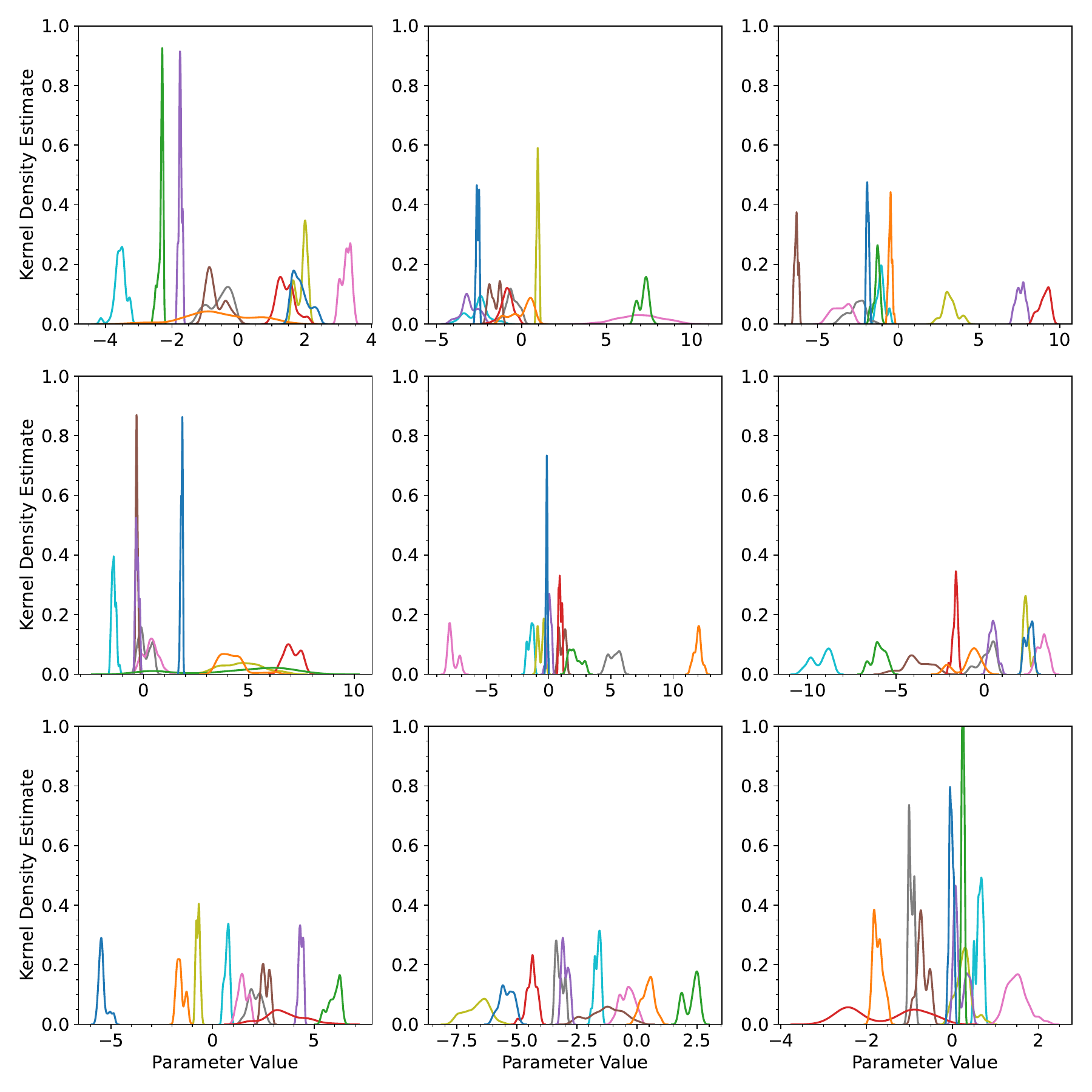}
    \caption{For the univariate cubic polynomial $f(x) = 1 + x + 2x^2 + 4x^3$, a third order Bayesian polynomial neural network was trained with the No-U-Turn-Sampler (NUTS) algorithm.  The kernel density estimates for the posterior distributions of the weights and biases of the polynomial neural network are shown. The panes are sequentially ordered (left-to-right, top-to-bottom) from the first layer to the last layer in the neural network.}
    \label{fig:Cubic_posteriors_nuts_w&b}
\end{figure}

For MCMC with NUTS, we used the Python library BlackJAX \cite{blackjax2020github} to perform sampling.  Since we had no prior knowledge of the weights and biases of the polynomial neural network but knew they weren't large values, we used the noninformative Gaussian prior with zero mean and standard deviation of 100.  The warmup was set to 500 steps and the number of steps taken following warmup was 500.  It took approximately 10 minutes for the code to run on a basic GPU.  The code can also execute on a CPU within practical timeframes (a few extra minutes over GPU execution time).  Since our neural network has a relatively small number of parameters, we plotted the kernel density estimates for the posterior distributions of the weights and biases of the polynomial neural network prior to expanding out the terms with Monte Carlo (see Figure \ref{fig:Cubic_posteriors_nuts_w&b}).  Most of the posterior distributions are close to being unimodal and symmetric, which initially suggests that the Laplace approximation and variational inference with a multivariate Gaussian should work towards estimating the posteriors.  The Laplace approximation approach takes significantly less time than MCMC - (1 minute vs 30 minutes).  Our results from the section and the following section provide enough evidence to use the Laplace approximation.  

Since this regression problem can also be posed as a Bayesian linear regression problem with a closed-form solution, we also solved it via simple Bayesian linear regression.  For Bayesian linear regression, we write the model as $ y = X B $.  The posterior distribution is defined by:

\begin{equation}
    w,b \sim \mathcal{N} \left (B=(X^T X)^{-1} X^T y, \frac{\beta^2}{(X^T X)} \right ),
\end{equation}

\noindent where the noise of the data ($\beta^2$) can be approximated by the sample variance of $(XB - y_{known})$.  This approach did not use any neural networks as its intended purpose was solely method validation.

\begin{figure}
    \centering
    \includegraphics[width=0.97\textwidth]{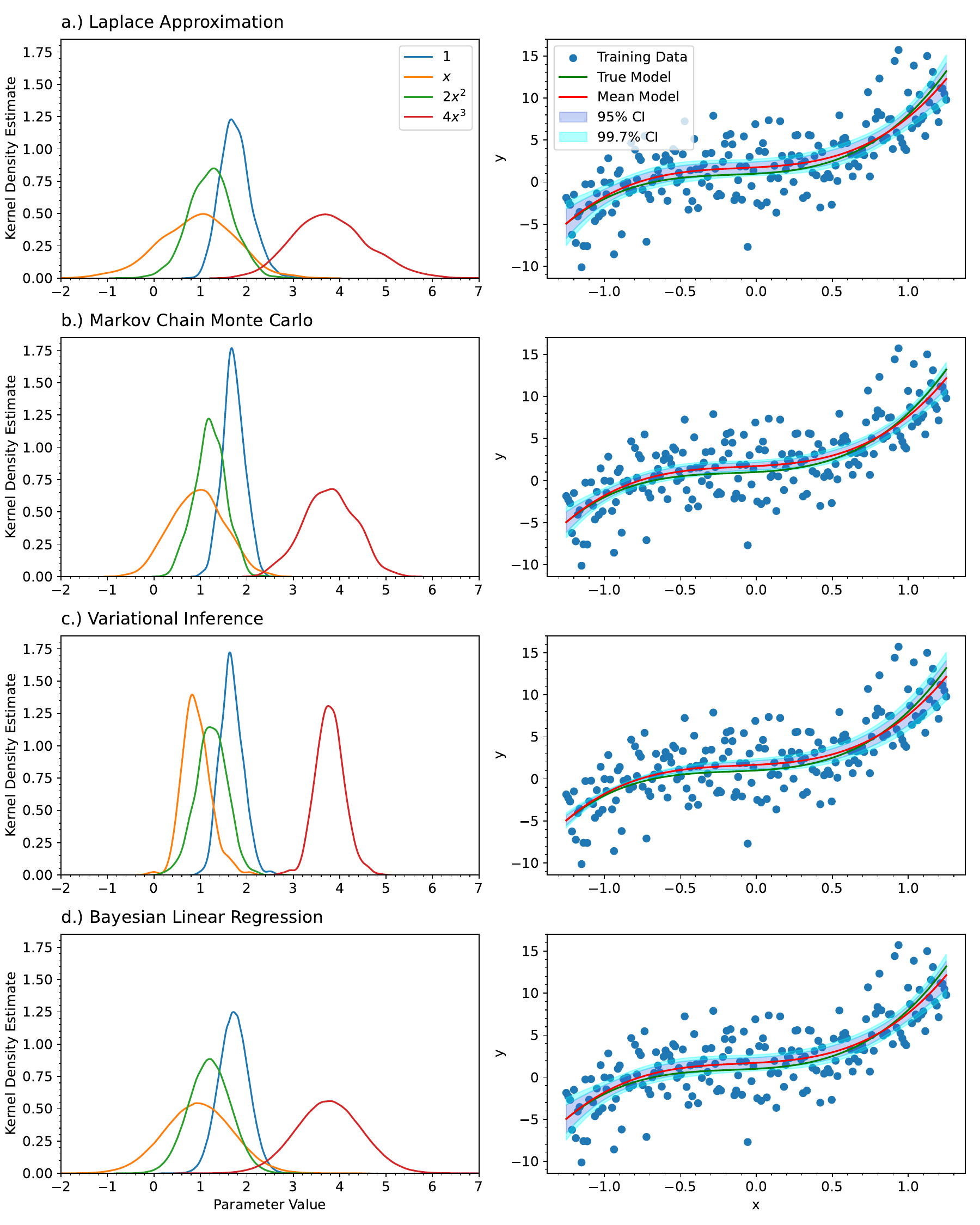}
    \caption{For the univariate cubic polynomial $f(x) = 1 + x + 2x^2 + 4x^3$, a third order Bayesian polynomial neural network was trained with a) the Laplace approximation, b) Markov Chain Monte Carlo with the No-U-Turn-Sampler (NUTS) algorithm, and c) Variational Inference.  For comparision, d) Bayesian linear regression was also performed on the training data. The kernel density estimates for the posterior distributions of the polynomial coefficients are shown (left) along with their predictions and confidence intervals (right).  For the left column, the true value of the parameters is shown in the legend.  Each of the columns share the same legend.}
    \label{fig:Cubic_Combined}
\end{figure}

In Figure \ref{fig:Cubic_Combined}, we show the kernel density estimates for the posterior distributions of the coefficients of the polynomial for the Laplace approximation, MCMC with NUTS, variational inference, and Bayesian linear regression.  Figure \ref{fig:Cubic_Combined} also shows the model predictions corresponding to the posterior distributions found by each of the methods along with 95\% and 99.7\% confidence intervals.  All of the methods have very similar results.  MCMC predicted slightly narrower posteriors than Bayesian linear regression, whereas the Laplace approximation predicted slightly wider posteriors; however, MCMC is the most computationally expensive of the methods and scales the worst as the number of model parameters increases.  Variational inference had the narrowest predictions for the posterior distributions, which resulted in a narrower confidence interval for the function evaluation.  Variational inference's confidence intervals were the only confidence intervals that failed to completely capture the true model data curve.  Variational inference was also comparatively difficult to train.  A notable amount of trial and error was required in order to guess plausible mean and covariance values to initialize the multivariate Gaussian approximation.  Different initial mean and covariance matrices worked for each problem and there is no hyperparameter optimization that can be performed to speed this up.  These problems will be addressed in future work.  However, variational inference is still computationally cheaper than MCMC.

\onecolumngrid
\clearpage
\FloatBarrier
\twocolumngrid

\onecolumngrid
\clearpage
\FloatBarrier
\twocolumngrid

\onecolumngrid
\clearpage
\begin{figure}
    \centering
    \includegraphics[width=1.0\textwidth]{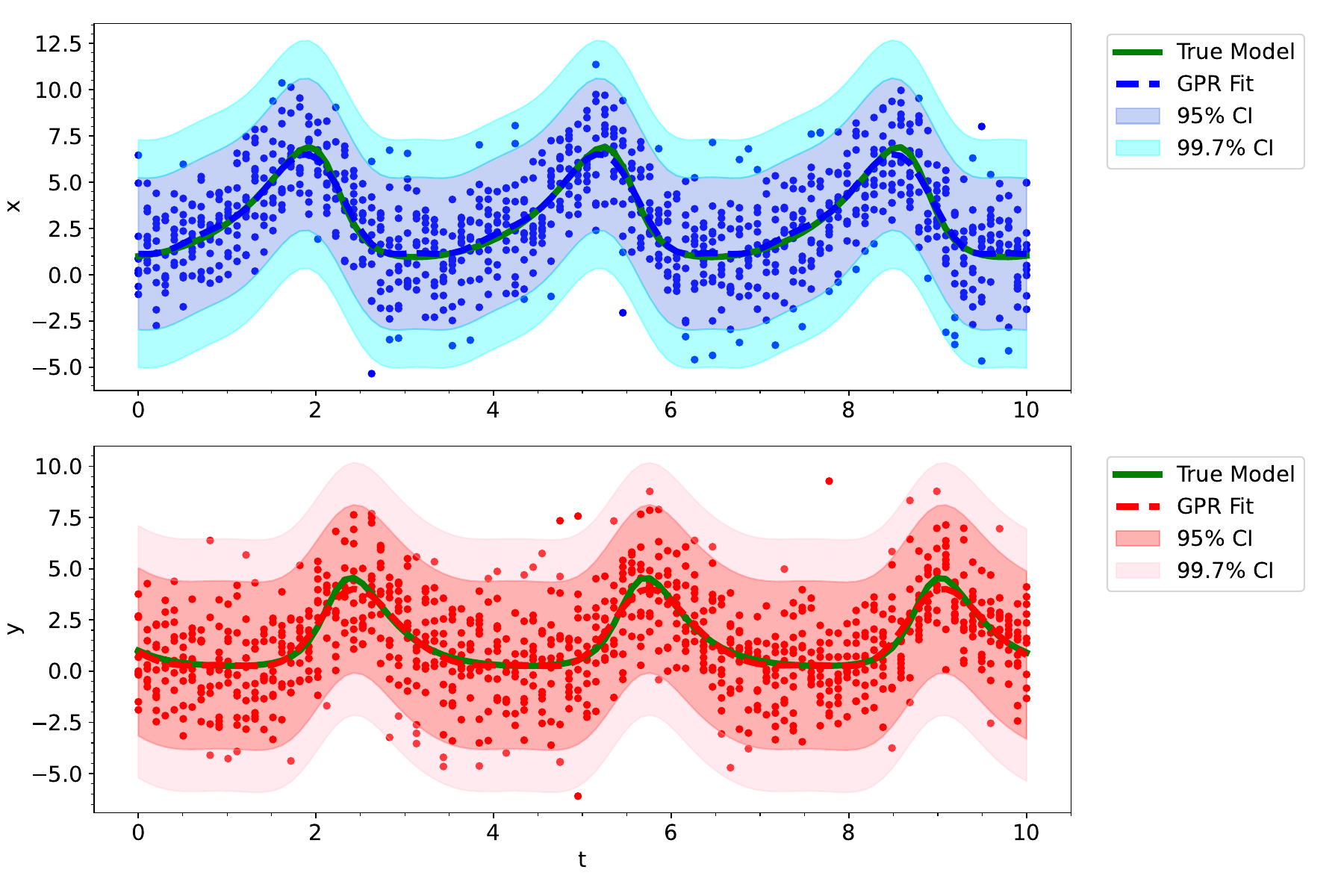}
    \caption{The fitted Gaussian process regression model trained on the noisy Lotka Volterra Oscillator data was used as initial conditions for the neural ODE's integration training trajectories.}
    \label{fig:Lotka_GPR_fit}
\end{figure}
\FloatBarrier
\twocolumngrid

\subsection{Lotka Volterra Deterministic Oscillator}

Our first demonstration of Bayesian parameter estimates for polynomial neural ODEs is on the deterministic Lotka-Volterra ODE model \citep{Lotka1925, volterra1926variazioni}, which describes predator-prey population dynamics, such as an ecosystem of rabbits and foxes.  When written as a set of first order nonlinear ODEs, the model is given by:

\begin{align} \label{eqn:lotka}
    \frac{dx}{dt} &= 1.5 x - x y, \\
    \frac{dy}{dt} &= -3 y + x y.
\end{align}

\noindent We generated our training data by integrating the initial value problem (IVP) with initial conditions $x=1$ and $y=1$ at $N=100$ points uniformly spaced in time between 0 and 10.  Since the Lotka-Volterra model is non-stiff, we used SciPy \cite{2020SciPy-NMeth} and DOPRI5 \cite{dopriref}, a fourth order accurate embedded method in the Runge–Kutta family of ODE solvers.  We then generated 10 high-noise trajectories originating from the same initial value by adding zero-centered Gaussian noise with a standard deviation of 2 to the training data.  This corresponds to a signal-to-noise ratio \cite{van2006introduction} (SNR or S/N) between 0.125 and 3.5.  See Figure \ref{fig:Lotka_GPR_fit}, ignoring the shaded GPR fit, to see the noisy training data.  The architecture from Ref.~\onlinecite{doi:10.1063/5.0130803} was used with 160 total parameters.  As discussed in the methods section, we batched our data into $N_t=89$ trajectories of consisting of 12 consecutive data points from the time series.  We simultaneously solved these batch trajectories during each epoch using our own JAX based differentiable ODE solver for the multistep fourth order explicit Runge–Kutta–Fehlberg method \citep{fehlberg1968classical}, which allows us to directly perform backpropagation through the ODE discretization scheme.

All of the Bayesian neural ODE approaches that we explored require integrating the neural ODE from starting initial conditions and comparing the prediction to the true data; however, since the data is extremely noisy, we cannot use the observed data points as initial conditions.  To generate good initial guesses, we used Gaussian process regression (GPR) on the noisy data prior to the model training process.  Since the Lotka-Volterra model is oscillatory, we used the Exp-Sine-Squared kernel \cite{duvenaud2014automatic} (also referred to as the periodic kernel), scaled by a constant kernel, along with the white kernel ($W_k$):

\begin{equation}
    k(x_i, x_j) = c^2 \exp \left (  - \frac{2 \sin^2(\pi d(x_i, x_j)/p)}{l^2} \right ) + W_k(\sigma_{GPR}^2).
\end{equation}

\noindent where $c^2$ is the constant for the constant kernel, $d$ is the euclidean distance function, $l$ is the length-scale, $p$ is the periodicity, and $\sigma_{GPR}^2$ is the variance of the Gaussian noise \cite{duvenaud2014automatic}.  We used MLE to obtain values for all of these unknown hyperparameters.  See Figure \ref{fig:Lotka_GPR_fit} for the GPR fit on the observed data.  

\begin{figure}
    \centering
    \includegraphics[width=1.0\textwidth]{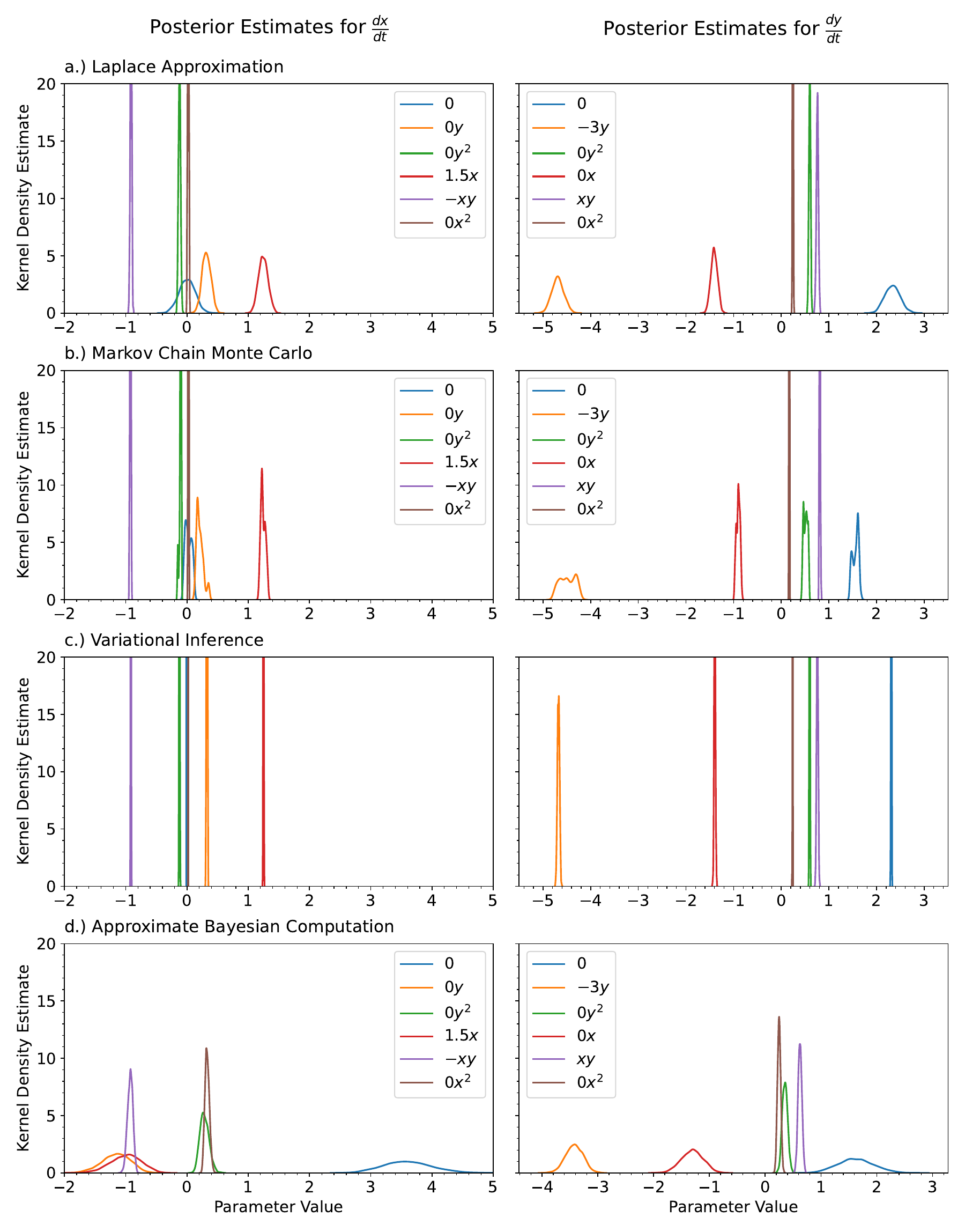}
    \caption{For the Lotka Volterra Oscillator, we show the kernel density estimates for the posterior distributions of the polynomial coefficients obtained with a.) the Laplace Approximation, b.) Markov Chain Monte Carlo, and c.) Variational Inference.  For comparison purposes, we also show the case for d.) Approximate Bayesian Computation on a normal ODE. The true value of the coefficients is shown in the legend.  The legend is shared for each of the columns.}
    \label{fig:Lotka_kde}
\end{figure}

In Figure \ref{fig:Lotka_kde}, we show the kernel density estimates for the posterior distributions of the ODE's parameters for the Laplace approximation, MCMC, and variational inference with a multivariate Gaussian approximation.  For comparison, we also show the inferred parameters obtained by vanilla Sequential Monte Carlo Approximate Bayesian Computation \cite{sisson2018handbook} (SMC ABC), a standard method used for inference of parameters in ODEs, for the ODE without any neural networks.  We wrote our own JAX based ABC method, but we recommend StochSS \cite{10.1371/journal.pcbi.1005220, StochSSLive, sciope} for those who'd like to use an existing toolkit. 
 SMC ABC had the worst performance for the true parameter values.  ABC predicted really wide posterior distributions for some of parameters that were far away from the true parameter values.  The Laplace approximation, Markov Chain Monte Carlo, and variational inference predicted more similar posterior distributions for the ODE parameters. As in the case for the univariate cubic polynomial, variational inference predicted very narrow posterior distributions.  MCMC resulted in very jagged posterior distributions.  

\begin{figure}
    \centering
    \includegraphics[width=0.95\textwidth]{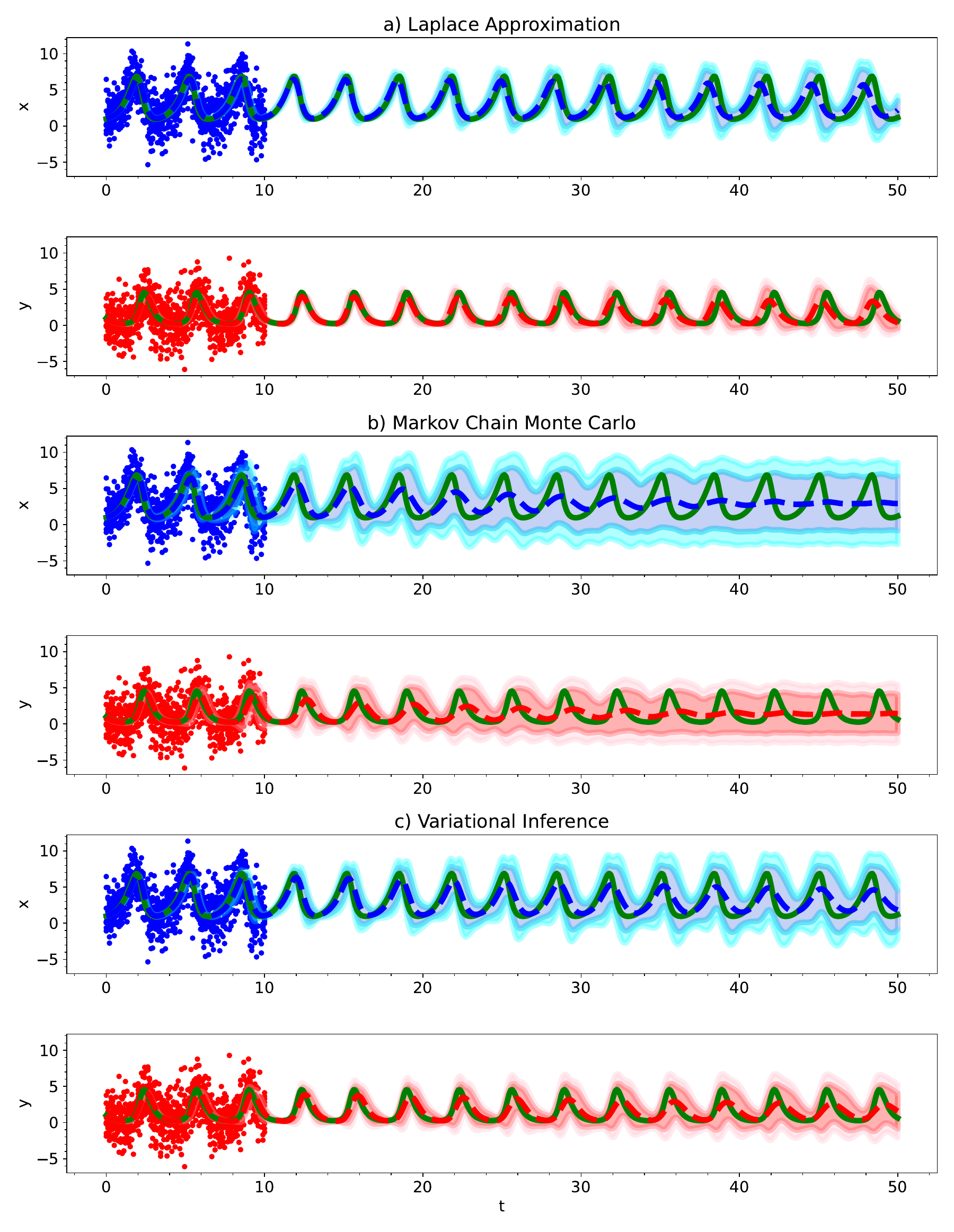}
    \caption{For the Lotka Volterra Oscillator, we show the predictive performance of a Bayesian polynomial neural ODE trained using a) the Laplace Approximation, b) Markov Chain Monte Carlo, and c) Variational Inference.  The solid red and blue dots indicate the training data, solid green lines indicate the true ODE model, dashed lines indicate the predictive mean model, and shaded regions indicate 95\% and 99.75\% confidence intervals.}
    \label{fig:Lotka_extrapolation}
\end{figure}

In Figure \ref{fig:Lotka_extrapolation}, we show the predictive performance of the inferred parameters.  For the parameters obtained from each of the methods, we integrated the ODE out to a final time 5 times that of the training data's time range.  The mean predicted model along with 95\% and 99.97\% confidence intervals is shown along with the training data and true ODE model used to generate the training data.  MCMC had the worst predictive performance; it predicted the oscillations to dampen over time.  Variational inference had reasonable performance with only minor dampening of the oscillations over time, but its predicted posteriors weren't ideal.  The Laplace approximation had the best predictive performance with only minor dampening and a minor phase shift, which further highlights its performance and usability since it is also the fastest and easiest method to train.  

\onecolumngrid
\clearpage
\FloatBarrier
\twocolumngrid

\onecolumngrid
\clearpage
\FloatBarrier
\twocolumngrid

\onecolumngrid
\clearpage
\begin{figure}
    \centering
\includegraphics[width=1.0\textwidth]{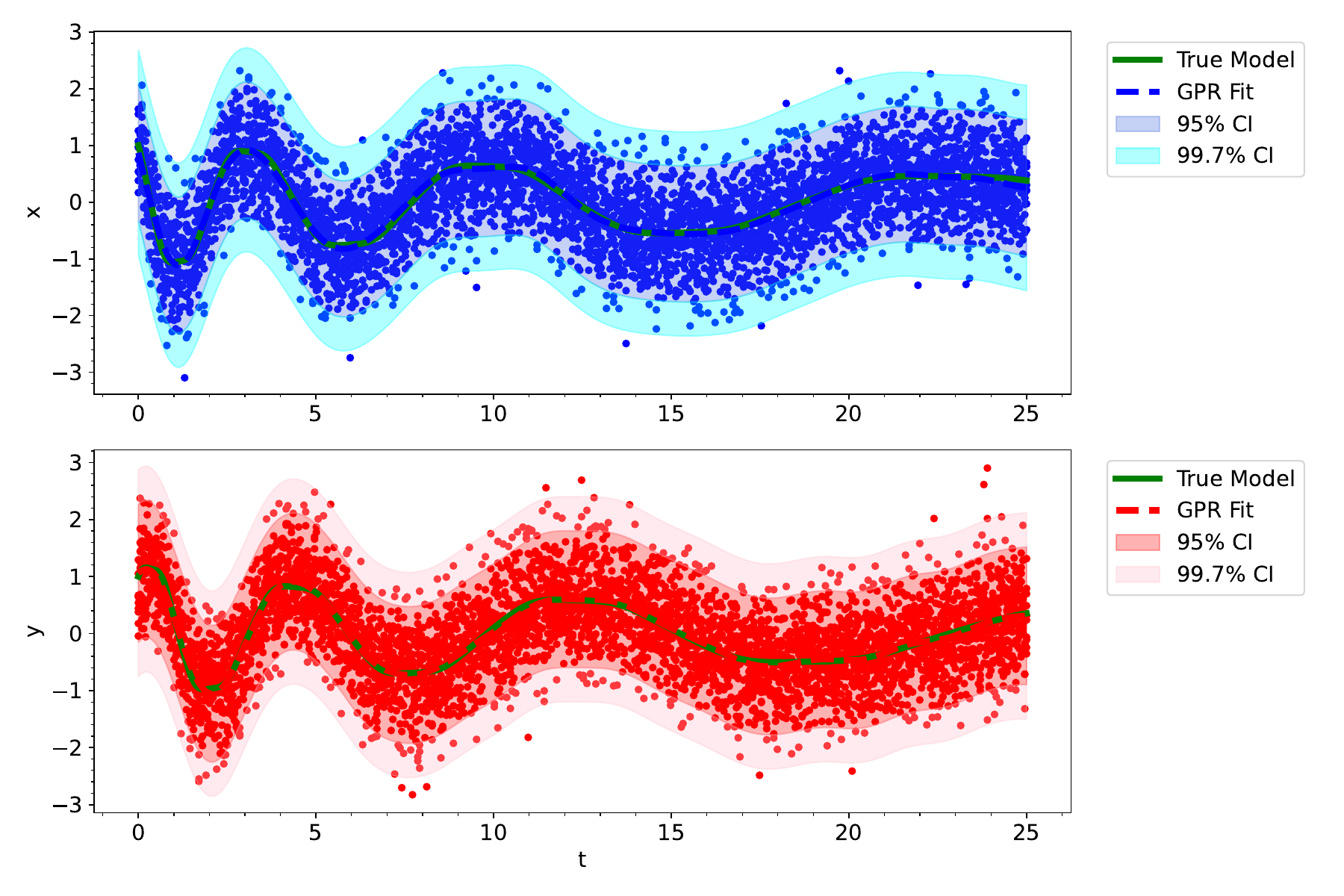}
    \caption{The fitted Gaussian process regression model trained on the noisy Damped Oscillator data was used as initial conditions for the neural ODE's integration training trajectories.}
    \label{fig:Damped_Oscillator_GPR_fit}
\end{figure}
\FloatBarrier
\twocolumngrid

\subsection{Damped Oscillatory System}

Our next example is the deterministic damped oscillatory system.  This model is a popular toy model for the field of neural ODEs \citep{chen2018neural, CollocationNeuralODE}.  Damped oscillations appear in many fields of biology, physics, and engineering \citep{doi:10.1080/00107514.2011.644441, karnopp1990system}.  One version of the damped oscillator model is given by:

\begin{flalign} \label{eqn:Damped_Oscillator}
    \frac{dx}{dt} &= -0.1 x^3 -2y^3 \\
    \frac{dy}{dt} &= 2x^3 -0.1y^3.
\end{flalign}

\noindent We generated our training data by integrating an initial value problem with initial conditions given by $(x_0, y_0) = (1,1)$ over the interval $t \in [0,25]$ for 500 points uniformly spaced in time.  Since the damped oscillator is also a nonstiff ODE system, we integrated the initial value problem with the same numerical methods as was done for the Lotka-Volterra model.  We then generated 10 high-noise trajectories originating from the same initial value by adding zero-centered Gaussian noise with a standard deviation of 0.6 to the training data.  This corresponds to an instantaneous signal-to-noise ratio \cite{van2006introduction} ranging from 0 to 2.1.  See Figure \ref{fig:Damped_Oscillator_GPR_fit}, ignoring the shaded GPR fit, to see the noisy training data.  

The architecture from Ref.~\onlinecite{doi:10.1063/5.0130803} was used with 660 total parameters.  For the training process, we created batches of trajectories consisting of 13 consecutive data points from the time series.  The number of consecutive points to include was determined by trial and error and unfortunately varies from model to model.  We simultaneously solve these batch trajectories during each epoch using our own JAX based differentiable ODE solver for the multistep fourth order explicit Runge–Kutta–Fehlberg method \citep{fehlberg1968classical}, which allows us to directly perform backpropagation through the ODE discretization scheme.  We have previously discussed why we chose this approach in the methods section of the paper.  

Prior to training our neural ODE, we used a smoothing algorithm to generate good initial values for our batch trajectories.  One can use any smoothing/filtering algorithm, but we used Gaussian process regression (GPR).  For this model, we had the best results with the use of a rational quadratic kernel \cite{duvenaud2014automatic} scaled by a constant kernel, along with a white kernel ($W_k$):

\begin{equation}
    k(x_i, x_j) = c^2  \left (1 + \frac{d(x_i, x_j)^2)}{2 \alpha l^2} \right )^{- \alpha} + W_k(\sigma_{GPR}^2).
    \label{eqn:rational_quadratic_GPR}
\end{equation}

\noindent where $c^2$ is the constant for the constant kernel, $d$ is the euclidean distance function, $l$ is the length-scale, $\alpha$ is the scale mixture parameter, and $\sigma_{GPR}^2$ is the variance of the Gaussian noise \cite{duvenaud2014automatic}.  We used MLE to obtain values for all of these unknown hyperparameters.  See Figure \ref{fig:Damped_Oscillator_GPR_fit} for the GPR fit on the observed data.  As you can see in the figure, the GPR model fits the noisy data extremely well.

\begin{figure}
    \centering
    \includegraphics[width=1.0\textwidth]{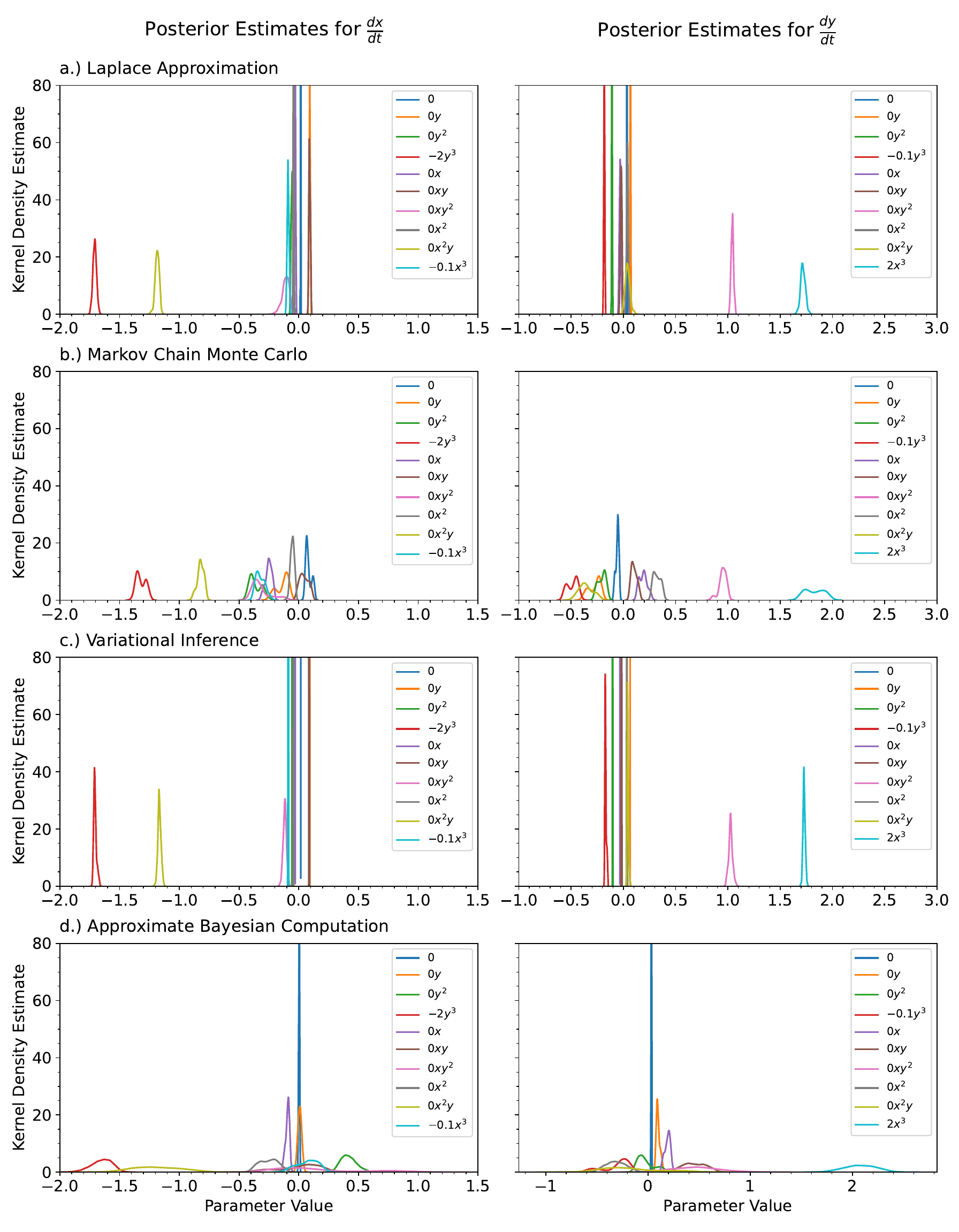}
    \caption{For the Damped Oscillator, we show the kernel density estimates for the posterior distributions of the polynomial coefficients obtained with a.) the Laplace Approximation, b.) Markov Chain Monte Carlo, and c.) Variational Inference.  For comparison purposes, we also show the case for d.) Approximate Bayesian Computation on a normal ODE. The true value of the coefficients is shown in the legend.  The legend is shared for each of the columns.}
    \label{fig:Damped_Oscillator_kde}
\end{figure}

In Figure \ref{fig:Damped_Oscillator_kde}, we show the kernel density estimates for the posterior distributions of the ODE's parameters for a) the Laplace approximation, b) Markov Chain Monte Carlo, and c) variational inference with a multivariate Gaussian approximation.  For comparison, we also show the inferred parameters obtained by vanilla Sequential Monte Carlo Approximate Bayesian Computation \cite{sisson2018handbook} (SMC ABC), a standard method used for inference of parameters in ODEs, for the ODE without any neural networks.  

\begin{figure}
    \centering
    \includegraphics[width=0.95\textwidth]{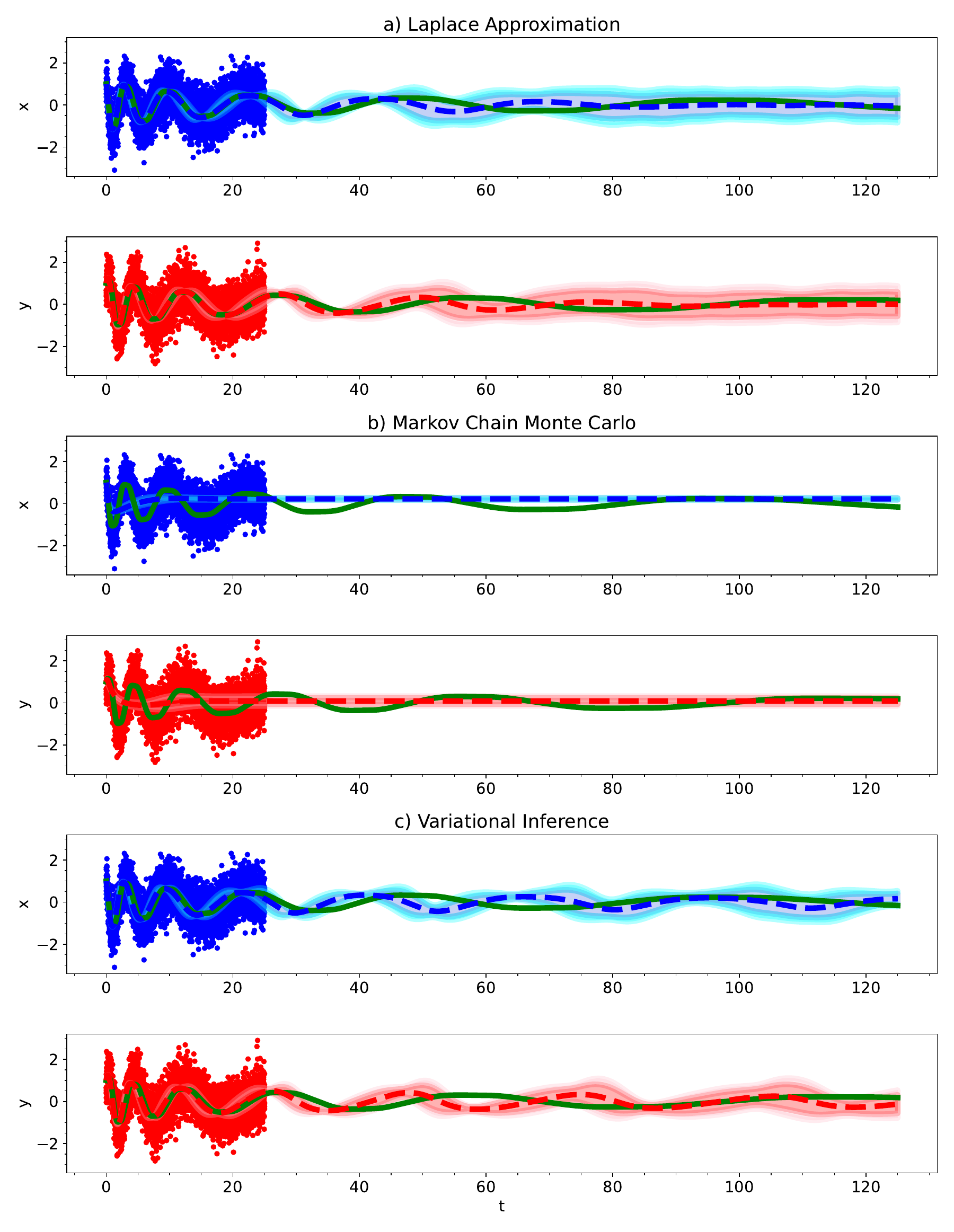}
    \caption{For the Damped Oscillator, we show the predictive performance of a Bayesian polynomial neural ODE trained using a) the Laplace Approximation, b) Markov Chain Monte Carlo, and c) Variational Inference.  The solid red and blue dots indicate the training data, solid green lines indicate the true ODE model, dashed lines indicate the predictive mean model, and shaded regions indicate 95\% and 99.75\% confidence intervals.}
    \label{fig:Damped_Oscillator_extrapolation}
\end{figure}

In Figure \ref{fig:Damped_Oscillator_extrapolation}, we integrated the final Bayesian models out to a final time 5 times that of the training data's time range.  The mean predicted model along with 95\% and 99.97\% confidence intervals is shown along with the training data and true ODE model used to generate the training data.

Generally speaking, we observed the same behavior for each of these methods as we did previously for the Lotka Volterra model.  Approximate Bayesian computation had the widest posterior distributions.  Variational inference had the narrowest posterior distributions and the confidence intervals in the trajectory prediction were too narrow to capture the true trajectory - the method is too confident about the inferred parameters.  This time, Markov Chain Monte Carlo (MCMC) completely failed to learn an accurate enough model to predict the trajectory of the system beyond $t=2$.  We spent a large amount of time playing around with the best settings for MCMC for this model, but the method failed every time.  The other methods did not require nearly as much time to get working results for. 
 Given enough patience, MCMC will result in somewhat accurate results but the other methods are much easier to use.  For this reason, we do not recommend using MCMC for neural ODEs.  The Laplace approximation provided the most accurate parameter estimates as well as predictions for the trajectories of the system.  It is also the fastest and easiest method to use.  For these reasons, we recommend using the Laplace approximation over the other methods.

\onecolumngrid
\clearpage
\FloatBarrier
\twocolumngrid

\onecolumngrid
\clearpage
\FloatBarrier
\twocolumngrid

\subsection{Lorenz Attractor}

The Lorenz attractor \cite{Lorenz1963DeterministicNF} is an example of a deterministic chaotic system \cite{ott2002chaos, hirsch2012differential} that came from a simplified model for atmospheric convection \cite{webster1999coupled}:   

\begin{flalign} \label{eqn:Lorenz}
    \frac{dx}{dt} &= \sigma (y-x), \\
    \frac{dy}{dt} &= x (r-z) -y, \\
    \frac{dz}{dt} &= xy - b z.
\end{flalign}

\noindent The equations describe the two-dimensional flow of a fluid with uniform depth between an upper and lower surface, given a temperature gradient.  In the equations, $x$ is proportional to the intensity of convective motion, $y$ is proportional to the difference in temperature between the rising and falling currents of fluid, and z is proportional to the amount of non-linearity within the vertical temperature profile \cite{Lorenz1963DeterministicNF, webster1999coupled}.  $\sigma$ is the Prandtl number, $r$ is the Rayleigh number, and $b$ is a geometric factor \cite{Lorenz1963DeterministicNF, webster1999coupled}.  Typically, $\sigma=10$, $r=28$, and $b=\frac{8}{3}$.  For our example, we use these values for the parameters.  Since the discovery of the Lorenz model, it has also been used as a simplified model for various other systems such as: chemical reactions \cite{Poland1993CooperativeCA}, lasers \cite{1975PhLA...53...77H}, electric circuits \cite{PhysRevLett.71.65}, brushless DC motors \cite{260218}, thermosyphons \cite{1981PhLA...82..439K}, and dynamos \cite{1981PhLA...82..439K}. 

Due to the chaotic nature of this system and the high frequency of oscillations, we required more training data for this example than for the previous examples shown.  We generated our training data from initial conditions $(x_0,y_0,z_0)=(1,1,1)$ over time interval $t \in [0,30]$ for 900 points uniformly spaced in time.  We then generated 10 high-noise trajectories originating from the same initial value by adding zero-centered Gaussian noise with a standard deviation of 2 to the training data.  The architecture from Ref.~\onlinecite{doi:10.1063/5.0130803} was used with 231 total parameters.  The training process was exactly the same as the previous two examples.  This time, we batched our data into training trajectories consisting of two adjacent data points.  This number was found through trial and error, but we hypothesize that the trajectory length needs to be shorter for this example due to the high frequency oscillations.  For this example, we used the same GPR kernel as was used for the damped oscillator (see Equation \ref{eqn:rational_quadratic_GPR}).

\begin{sidewaysfigure}
    \centering
    \includegraphics[width=0.95\textwidth]{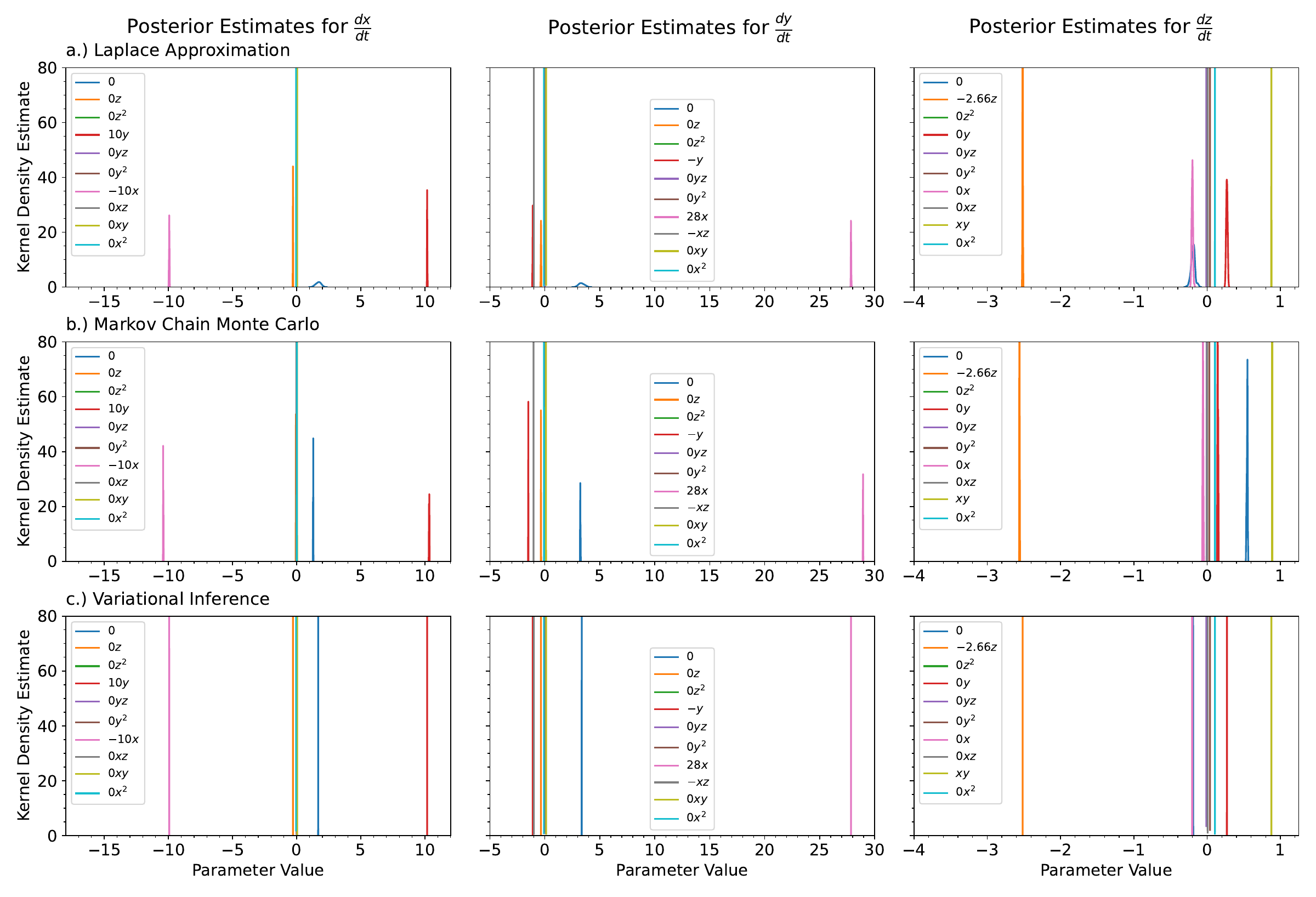}
    \caption{For the Lorenz Attractor, we show the kernel density estimates for the posterior distributions of the polynomial coefficients obtained with a.) the Laplace Approximation, b.) Markov Chain Monte Carlo, and c.) Variational Inference. The true value of the coefficients is shown in the legend. The legend is shared for each of the columns.
}
    \label{fig:Lorenz_kde}
\end{sidewaysfigure}

\begin{sidewaysfigure}
    \centering
    \includegraphics[width=0.95\textwidth]{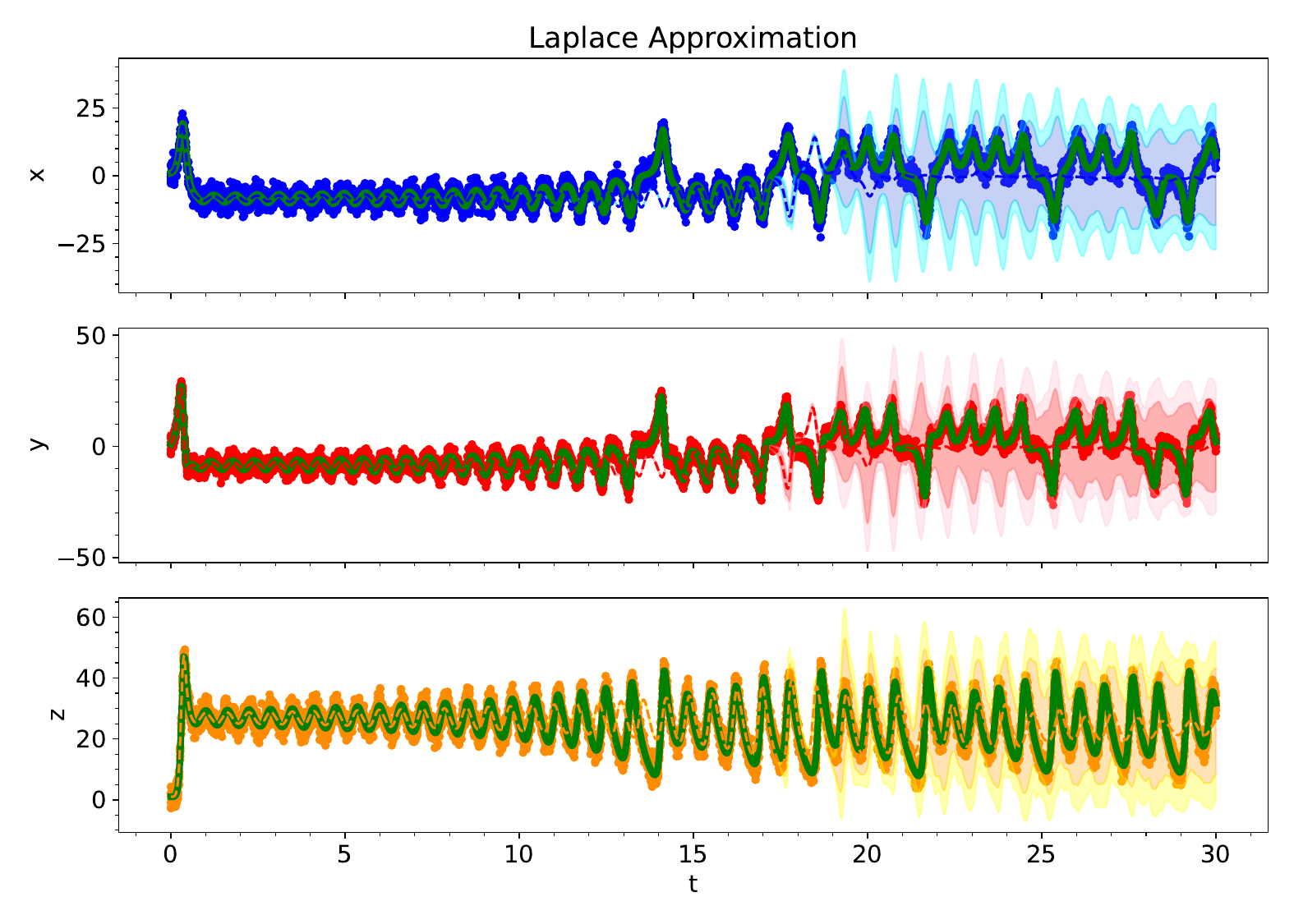}
    \caption{For the Lorenz Attractor, we show the predictive performance of a Bayesian polynomial neural ODE trained using the Laplace Approximation.  The solid red, blue, and orange dots indicate the training data, solid green lines indicate the true ODE model, dashed lines indicate the predictive mean model, and shaded regions indicate 95\% and 99.75\% confidence intervals.}
    \label{fig:Laplace_Lorenz_extrapolation}
\end{sidewaysfigure}

\begin{sidewaysfigure}
    \centering
    \includegraphics[width=0.95\textwidth]{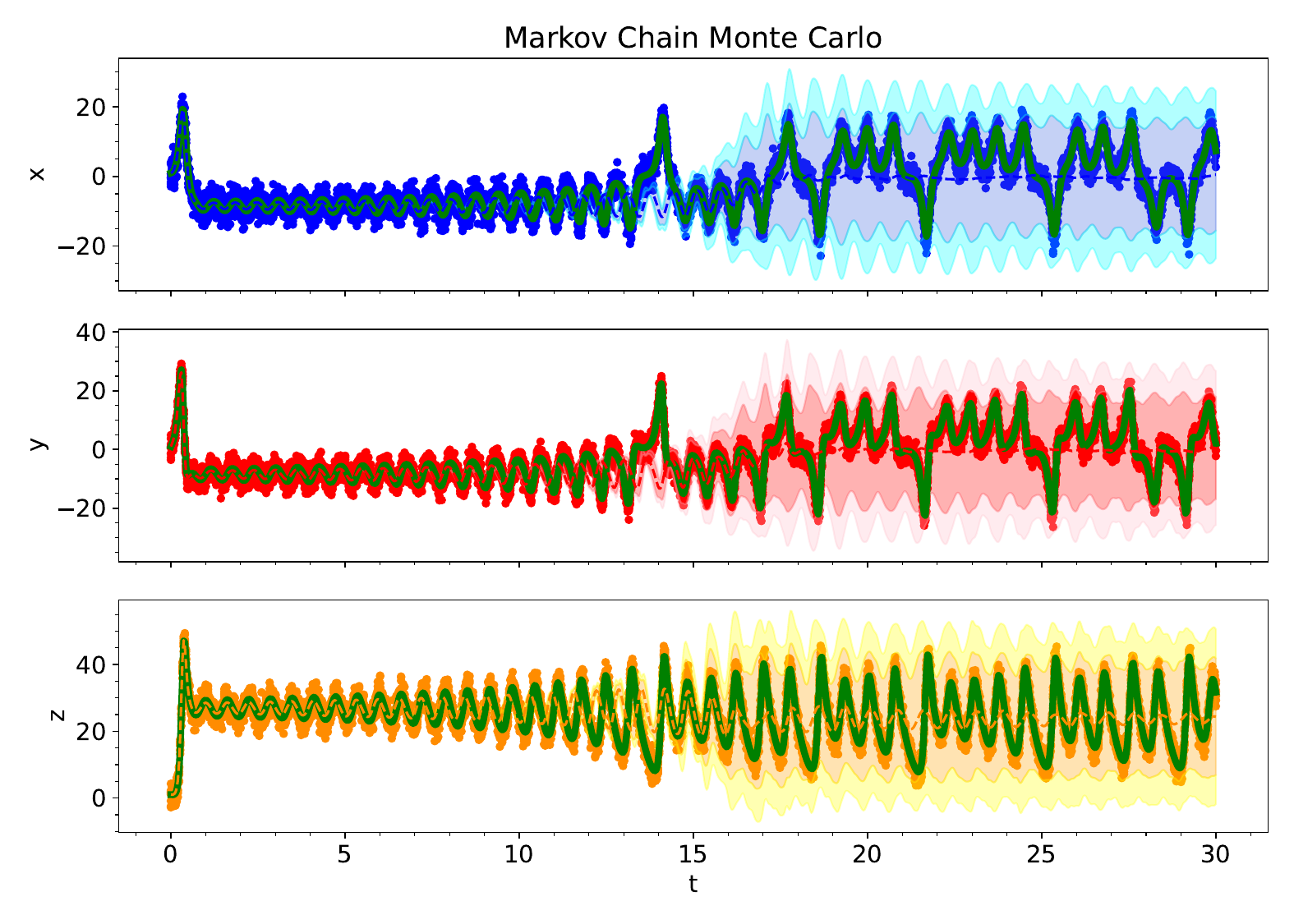}
    \caption{For the Lorenz Attractor, we show the predictive performance of a Bayesian polynomial neural ODE trained using Markov Chain Monte Carlo.  The solid red, blue, and orange dots indicate the training data, solid green lines indicate the true ODE model, dashed lines indicate the predictive mean model, and shaded regions indicate 95\% and 99.75\% confidence intervals.}
    \label{fig:MCMC_Lorenz_extrapolation}
\end{sidewaysfigure}

\begin{sidewaysfigure}
    \centering
    \includegraphics[width=0.95\textwidth]{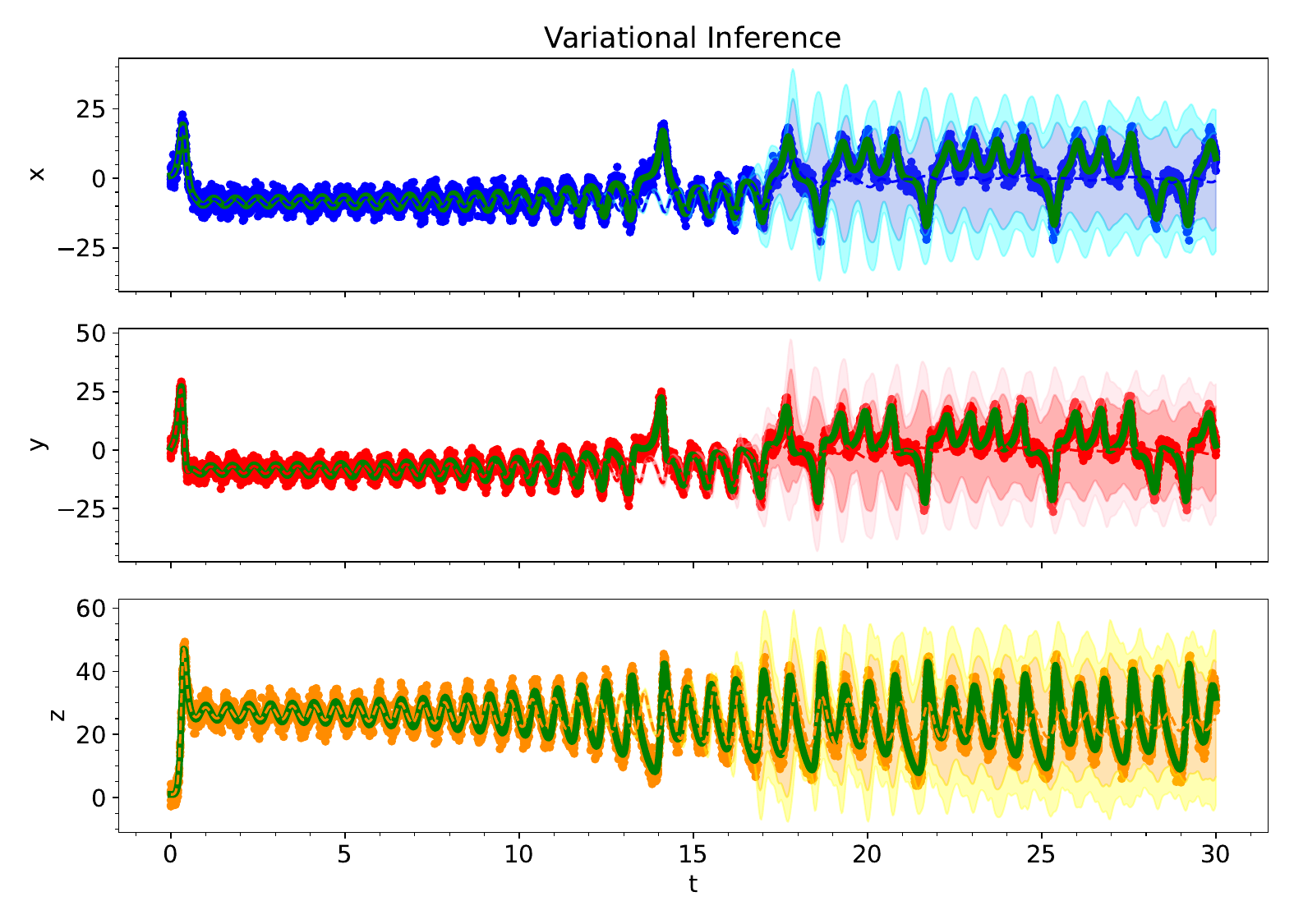}
    \caption{For the Lorenz Attractor, we show the predictive performance of a Bayesian polynomial neural ODE trained using variational inference.  The solid red, blue, and orange dots indicate the training data, solid green lines indicate the true ODE model, dashed lines indicate the predictive mean model, and shaded regions indicate 95\% and 99.75\% confidence intervals.}
    \label{fig:VI_Lorenz_extrapolation}
\end{sidewaysfigure}

Figure \ref{fig:Lorenz_kde} shows the kernel density estimates for the various Bayesian inference methods.  Since the level of noise is smaller for this example, the posterior estimates are narrower.  There is also very little difference between the predicted Bayesian uncertainties.  Figures \ref{fig:Laplace_Lorenz_extrapolation}, \ref{fig:MCMC_Lorenz_extrapolation}, and \ref{fig:VI_Lorenz_extrapolation} show the trajectory predictions for the Laplace approximation, Markov Chain Monte Carlo, and variational inference respectively.  All of the methods' 95\% confidence intervals were able to capture the true trajectory.  In terms of accuracy, there is no clear winner for the Lorenz attractor.  However, in terms of speed, the Laplace approximation is the best choice.

\onecolumngrid
\clearpage
\FloatBarrier
\twocolumngrid

\onecolumngrid
\clearpage
\FloatBarrier
\twocolumngrid

\onecolumngrid
\clearpage
\FloatBarrier
\twocolumngrid

\onecolumngrid
\clearpage
\FloatBarrier
\twocolumngrid

\newpage
\clearpage
\subsection{Learning Missing Terms from a Partially Known ODE Model}

It is common for scientists to have an incomplete model of their system - one in which they are confident about certain processes undergoing the system, but there are mechanisms they aren't aware of.  Rather than learn the whole system from scratch, we can incorporate the known parts of our model into the neural ODE and have the neural ODE suggest additional components of the ODE model given the observed data.  Incorporating the known ODE model into the neural ODE framework is done simply by adding the output of the known equation to that of the neural ODE output - no special treatment is required aside from that (see Equation \ref{eqn:missing_terms}).  

We will use the Lotka Volterra Oscillator to demonstrate the ability of polynomial neural ODEs to learn missing terms from a partially known ODE model.  We also show that Bayesian uncertainties can be obtained for the parameters in the terms that the neural ODE suggests including.  As a reminder, the Lotka Volterra model is given by:

\begin{align} \label{eqn:lotka-missing}
    \frac{dx}{dt} &= \colorbox{yellow}{\textbf{1.5 x}} - x y, \\
    \frac{dy}{dt} &= -3 y + \colorbox{yellow}{\textbf{x y}}.
\end{align}

\noindent For this experiment, the highlighted terms are the ones we do not know.  The goal will be to recover these terms along with posterior distributions for the values of the parameters.  We used the same training data, GPR model for the initial conditions, and training process as was previously used in the Lotka Volterra example.  The only difference was including the known ODE model (see Equation \ref{eqn:missing_terms}).

\begin{figure}
    \centering
    \includegraphics[width=0.95\textwidth]{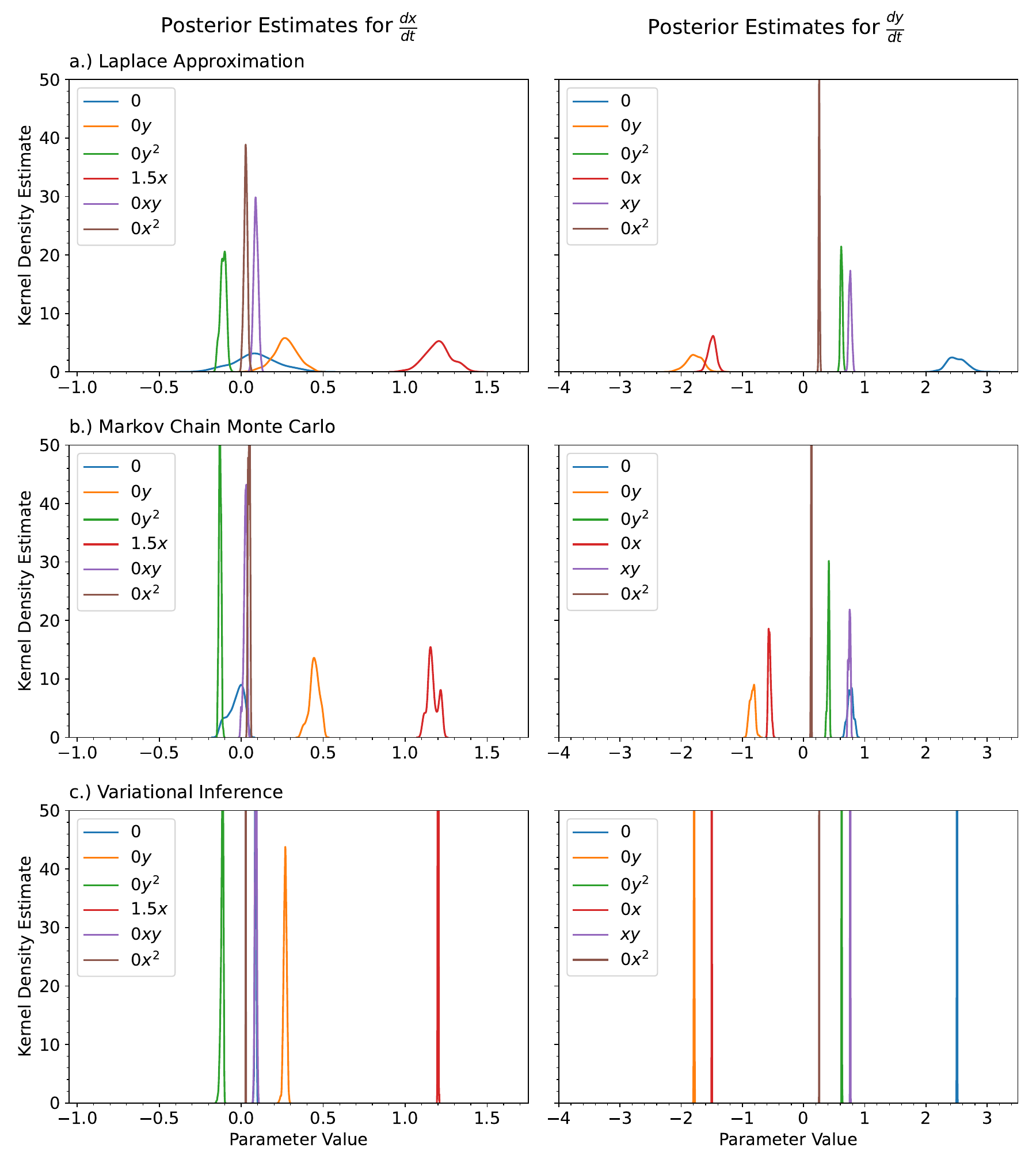}
    \caption{It is common for a domain expert to understand part of the system's underlying mechanisms, but have in incomplete model.  Given an incomplete model, a neural ODE can learn the missing terms from the ODE model that best fit the observed data.  We have removed two of the terms from the Lotka Volterra model and tested the neural ODE's ability to learn the missing terms.  We show the kernel density estimates for the posterior distributions of the polynomial coefficients obtained with a.) the Laplace Approximation, b.) Markov Chain Monte Carlo, and c.) Variational Inference. The true value of the coefficients is shown in the legend. The legend is shared for each of the columns.
}
    \label{fig:Lotka_missing_kde}
\end{figure}

Figure \ref{fig:Lotka_missing_kde} shows the posterior distributions recovered for all of the candidate terms to include in the final ODE model.  The neural ODE was able to identify the missing terms with few false terms.  Most of the terms that are not in the true model are predicted to be close to zero.  As was seen in the previous examples, variational inference provides very narrow posterior distributions and MCMC provides results between the Laplace approximation and variational inference.  

\section{Conclusion}

This work addressed the problem of how to handle noisy data and recover uncertainty estimates for: (1) symbolic regression with deep polynomial neural networks and (2) polynomial neural ODEs.  More broadly, we also helped to answer the question of how to handle noisy data and perform Bayesian inference on the general class of symbolic neural networks and symbolic neural ODEs.  

We compared the following Bayesian inference methods: (a) the Laplace approximation, (b) Markov Chain Monte Carlo (MCMC) sampling methods, and (c) variational inference.  We do not recommend using Markov Chain Monte Carlo for neural ODEs.  Using MCMC for neural ODEs requires a substantial amount of patience, it is the most computationally expensive method, and we showed that the results are not encouraging.  A substantial amount of development work needs to be devoted towards addressing the challenges of using MCMC for neural ODEs in an effective manner.  Variational inference is also challenging to use - some time is spent deciding the mean and covariance matrix to use for initialization of the parameters.  This process can be sped up by first obtaining point estimates for the parameters and using the values obtained to initialize the mean matrix.  Using this approach made variational inference a viable option to implement.  However, the posterior estimates generated by variational inference's posterior are consistently too narrow: it is too confident about its estimates.  

The Laplace approximation is the easiest to implement and the fastest method.  The main challenge associated with the Laplace approximation for neural networks is inverting the Fisher information matrix; however, most of the models in this class of problems are small enough that this is not an issue.  Based on our experience, we recommend having no more than 50,000 parameters if you plan on using the Laplace approximation for a neural network and want to use the exact or pseudo inverse of the Fisher information matrix.  We were initially skeptical about the Laplace approximation because it makes a Gaussian approximation for all of the parameters.  However, we have shown that this approximation is not problematic when the polynomials are multiplied out.  We have shown that the Laplace approximation has high accuracy.  For these reasons, we recommend using the Laplace approximation for this class of problems.  


\onecolumngrid
\clearpage
\FloatBarrier
\twocolumngrid

\section{Acknowledgements}

The authors acknowledge research funding from NIBIB Award No. 2-R01-EB014877-04A1 (grant 2-R01-EB014877-04A1 to LRP). This work has benefited from our participation in Dagstuhl Seminar 22332 "Differential Equations and Continuous-Time Deep Learning \cite{dagstuhl}."  Many thanks to the organizers of this seminar: David Duvenaud (University of Toronto, CA), Markus Heinonen (Aalto University, FI), Michael Tiemann (Robert Bosch GmbH - Renningen, DE), and Max Welling (University of Amsterdam, NL).  This work has also benefited from our participation in the University of Bonn's Hausdorff School: “Inverse problems for multi-scale models.”  Many thanks to the organizers of this summer school: Lorenzo Contento, Jan Hasenauer, and Yannik Schälte.  We'd like to thank Alexander Franks (UC Santa Barbara) for giving us the idea of using Monte Carlo for obtaining posterior distributions for the polynomial coefficients.  We'd also like to thank Michael Tiemann and Katharina Ott (Robert Bosch GmbH - Renningen, DE) for recommending that we try the Laplace approximation on the polynomial neural ODEs.  Lastly, we'd like to thank Patrick Kidger (Google) for recommending that we switch from PyTorch to JAX and providing resources for making the switch, as JAX has allowed us to do so much more.

Use was made of computational facilities purchased with funds from the National Science Foundation (CNS-1725797) and administered by the Center for Scientific Computing (CSC). The CSC is supported by the California NanoSystems Institute and the Materials Research Science and Engineering Center (MRSEC; NSF DMR 2308708) at UC Santa Barbara.

This work was supported in part by National Science Foundation (NSF) awards CNS-1730158, ACI-1540112, ACI-1541349, OAC-1826967, OAC-2112167, CNS-2100237, CNS-2120019, the University of California Office of the President, and the University of California San Diego's California Institute for Telecommunications and Information Technology/Qualcomm Institute. Thanks to CENIC for the 100Gbps networks.

The content of the information does not necessarily reflect the position or the policy of the funding agencies, and no official endorsement should be inferred.  The funders had no role in study design, data collection and analysis, decision to publish, or preparation of the manuscript.

\nocite{*}
\bibliography{aipsamp}


\end{document}